\documentclass{article}

\newcommand{\CEG}{\mathrm{CEG}}
\newcommand{\CEL}{\mathrm{CEL}}
  \usepackage{bm}

 \usepackage{graphicx}
\PassOptionsToPackage{numbers, compress}{natbib}

 \usepackage[preprint]{neurips_2026}

\usepackage[utf8]{inputenc} 
\usepackage[T1]{fontenc}    
\usepackage{hyperref}       
\usepackage{url}            
\usepackage{booktabs}       
\usepackage{amsfonts}       
\usepackage{nicefrac}       

\usepackage{caption}

\usepackage{silence}
\usepackage{microtype}

\usepackage{xcolor}         
\usepackage{amsmath,amssymb}

\title{Internal Data Repetition Destroys Language Models}

\newcommand{\Dr}{D_{\mathrm{r}}}
\newcommand{\OT}{\mathrm{OT}}
\newcommand{\Rep}{R}
\newcommand{\flops}{C}
\newcommand{\baseline}{\mathrm{base}}

\author{%
\makebox[\textwidth][c]{%
\begin{tabular}{c}
\textbf{Jessica Chudnovsky}\textsuperscript{*\,1} \quad
\textbf{Joshua Kazdan}\textsuperscript{*\,2} \quad
\textbf{Noam Levi}\textsuperscript{*\,3} \quad
\textbf{Rylan Schaeffer}\textsuperscript{1} \\[0.25em]
\textbf{Yegor Denisov-Blanch}\textsuperscript{1} \quad
\textbf{Bo He}\textsuperscript{4} \quad
\textbf{Mehmet Donmez}\textsuperscript{4} \\[0.25em]
\textbf{Sanmi Koyejo}\textsuperscript{\textdagger\,1} \quad
\textbf{David Donoho}\textsuperscript{\textdagger\,2}
\end{tabular}%
}
}

\begin{document}

\maketitle


\begingroup
\renewcommand{\thefootnote}{\fnsymbol{footnote}}
\renewcommand{\theHfootnote}{authornote}
\footnotetext[1]{%
Equal contribution.
\textsuperscript{\textdagger}Senior author.
\textsuperscript{1}Stanford Computer Science.
\textsuperscript{2}Stanford Statistics.
\textsuperscript{3}Tel Aviv University.
\textsuperscript{4}IMC Trading.
Correspondence: \href{mailto:jchud@cs.stanford.edu}{\nolinkurl{jchud@cs.stanford.edu}}.
}
\endgroup

\begin{abstract}


Language models are running out of high-quality training data, and even aggressively deduplicated corpora retain some amount of repetition. Earlier controlled studies predated Chinchilla-style scaling laws and could only measure the cost of repetition indirectly. We revisit repetition in the Chinchilla era, using a fitted no-repetition scaling law to report Compute-Equivalent Gain and Compute-Equivalent Loss. We show that under this modernized paradigm, repetition damage is systematic in three ways. First, holding compute allocated to repeated data constant, eval loss peaks at an intermediate repeat count $\Rep$; repeating a moderately sized subset a moderate number of times damages performance more than repeating a large subset a few times or a small subset many times. Second, the location of this peak is well-fit by a power law in model size; this scaling law reveals that the most damaging number of repeated data grows more quickly than compute. Finally, when repeated documents consume 10\% of the FLOPs budget in a controlled exact-document repetition setting, the compute-equivalent loss can be large: on FineWeb-Edu-Dedup, the most damaging repeat count for a Qwen3-style 344M-parameter model at $\OT=1$ matches the loss of a no-repetition run using 67\% of the FLOPs.
We demonstrate that these phenomena are not language-model-specific, and can be analytically understood in a simple statistical model: a misspecified linear regression with verbatim duplicates reproduces the same qualitative loss peak, quantifying how such peaks can arise from a statistical tradeoff between memorization and generalization.
Our findings add precision to the study of duplication in language models, allowing practitioners to quantify the wasted compute incurred by the presence and repeat structure of duplicates in pretraining corpora.

\end{abstract}

\section{Introduction}

Pretraining has entered a data-constrained regime. The high-quality public text corpora used for frontier training have been exhausted~\citep{villalobos2024position,longpre2024pretrainers}, forcing multi-epoch training.  To maintain data efficiency in pretraining, flagship corpora such as FineWeb-Edu, DataComp-LM, Dolma, and RedPajama-v2~\citep{penedo2024fineweb,li2024datacomp,soldaini2024dolma,weber2024redpajama} have implemented aggressive deduplication and filtering. Yet aggressive deduplication is not perfect deduplication~\citep{lee2022deduplicating,abbas2023semdedup,tirumala2023d4}: pretraining streams retain near-duplicate documents, paraphrased templates, and semantically redundant web pages, and as scale grows the meaning of ``duplicate'' itself shifts~\citep{kazdan2026scale}. In this paper, we isolate one controlled case: exact document-level replay of a selected repeated pool. We precisely quantify the cost incurred by exact repeats during pretraining.

The closest earlier controlled study, \citet{hernandez2022scaling}, established that repeating a small training-data fraction non-monotonically degrades held-out loss and framed the damage as a reduction in \emph{effective parameter count}, an outdated measurement that tried to quantify performance in terms of parameters rather than compute, where all models were trained on 300B tokens irrespective of parameter count. This left the larger models undertrained while the smaller ones were grossly overtrained. \citet{hernandez2022scaling}'s work predated the widespread use of compute-optimal scaling~\citep{hoffmann2022training}, which provided a clean compute axis for scaling-law comparisons. Modern practitioners measure the cost of training in FLOPs, not parameters. We modernize the study of memorization scaling laws by measuring damage via \emph{Compute-Equivalent Gain ($\CEG$)}, following the compute-equivalent gains framework of \citet{gundlach2025originalgorithmicprogressai, davidson2023aicapabilitiessignificantlyimproved}, and \citet{musespark2026meta}. For a repeated-data run with loss $L$ and actual compute $C_{\mathrm{actual}}$, $\CEG$ is the no-repetition compute required to reach $L$ divided by $C_{\mathrm{actual}}$. We define \emph{Compute-Equivalent Loss} as $\CEL = 1-\CEG$. Thus, $\CEG=1$ matches the no-repetition baseline, while $\CEG<1$ indicates compute-equivalent loss.

To modernize the study of repetition damage, we train Qwen3-style models~\citep{yang2025qwen3} with $N \in \{34, 48, 63, 93, 153,\allowbreak 344\}\mathrm{M}$ parameters on FineWeb-Edu-Dedup~\citep{penedo2024fineweb}. We sweep the overtraining multiplier $\OT \in \{0.25, 0.5, 1, 2, 4\}$ and the per-document repeat count $\Rep$ on an approximately logarithmic grid from no repeats to as high as $\Rep=20000$ (we plot up to $\Rep \approx 3000$). Throughout, we fix the repeated-token fraction at $f=0.1$: 90\% of training tokens come from non-repeated documents, and the remaining 10\% come from a smaller document pool replayed $\Rep$ times. This fraction is large enough to produce measurable damage while keeping the bulk of training on unique data, and matches the repeated-token fraction used by \citet{hernandez2022scaling}, enabling a direct comparison.

The Chinchilla budget identity $C \approx 6NT = 120 \cdot \OT \cdot N^2$ ties model size, total tokens $T$, and total compute $C$ together~\citep{gadre2024language,porian2024resolving,kaplan2020scaling,sardana2024beyond}. This lets us vary repetition structure inside an otherwise fixed training budget, separating the effect of repeat concentration from total compute. For example, fixing the memorization compute budget at 10\%, we can repeat either $1\%$ of the pretraining corpus $10$ times, or we can repeat $0.01\%$ of the pretraining corpus $1000$ times. We refer to this pairing of repeated-pool size and repeat count as a \emph{repetition structure}.  We find that the most damaging repetition structure is a function of the model's parameter count $N$, and we translate the resulting loss increases into Compute-Equivalent Gain and Compute-Equivalent Loss using a fitted no-repetition scaling law. Under this fitted frontier, the most damaging repeat count at our largest scale has $\CEG\approx0.67$. A misspecified linear regression with verbatim duplicates reproduces the same qualitative non-monotonicity in closed form, suggesting that such peaks can arise from the statistical structure of duplicated samples rather than from any attention, depth, or optimizer-specific effects particular to language model training.

\begin{figure}[t]
\centering
\includegraphics[width=0.5\linewidth]{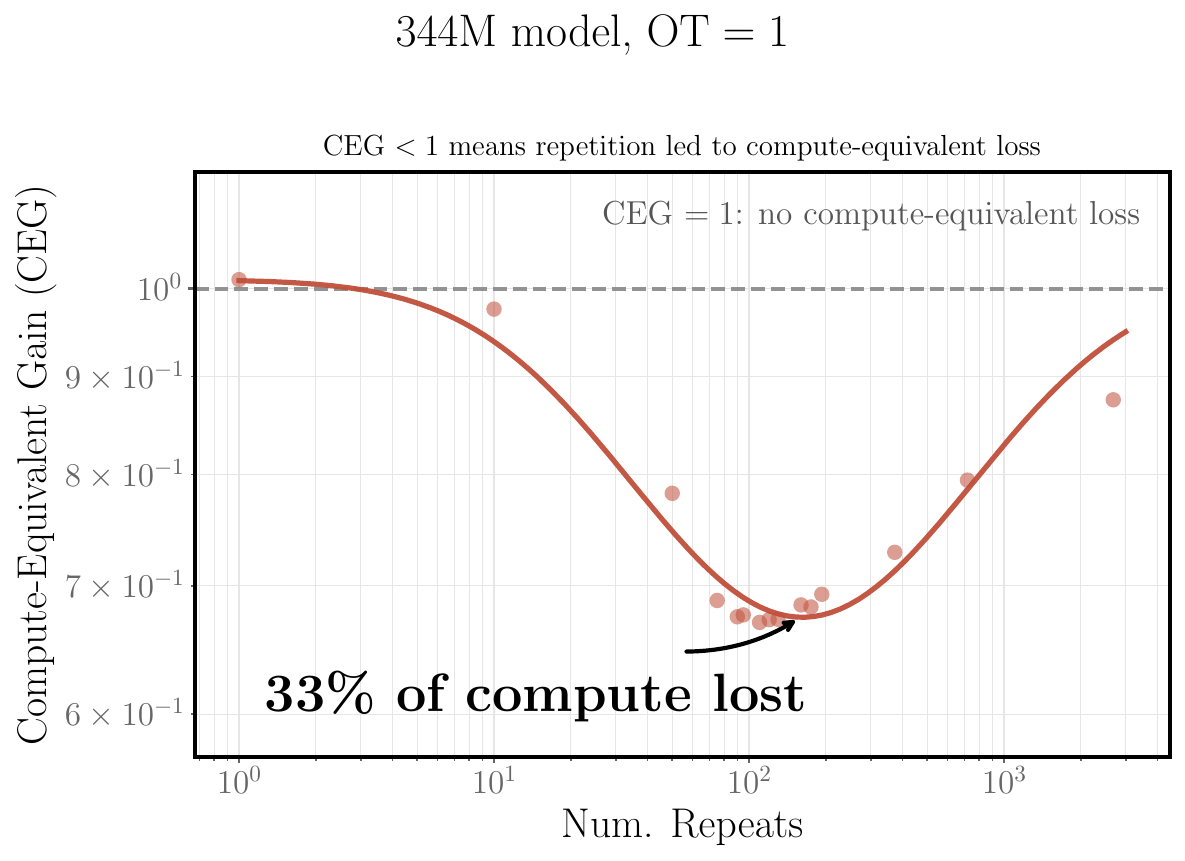}
\caption{
Compute-Equivalent Gain ($\CEG$) as a function of the per-document repeat count $R$ at fixed training compute for the Qwen3-style 344M-parameter model trained on FineWeb-Edu-Dedup at the Chinchilla-optimal multiplier $\OT = 1$. $\CEG$ is the ratio of the no-repetition compute that would reach the achieved loss to the compute actually spent. $\CEG = 1$ is when there is no gain or loss relative to the no repetition baseline. At $R$ near $100$, $\mathrm{CEL}$ rises to roughly $0.33$, meaning the run reaches the loss of a run without repetitions trained with only two-thirds of the FLOPs \cite{yang2025qwen3, penedo2024fineweb}.
}
\label{fig:teaser}
\end{figure}

\paragraph{Contributions.}
(i)  We quantify repeated-data damage in compute-equivalent units using a fitted no-repetition Chinchilla scaling law (\S\ref{sec:finding-3}). At 344M parameters and $\OT=1$ on FineWeb-Edu-Dedup \cite{yang2025qwen3, penedo2024fineweb}, the most damaging repeat count has $\CEL\approx0.33$. (ii) We identify the intermediate-$R$ regime where damage peaks (\S\ref{sec:finding-1}) and show that its location follows the power law $R^{\mathrm{peak}} \propto N^{-0.96}$ (\S\ref{sec:finding-2}), making the worst-case configuration predictable using $N$. (iii) We derive closed-form train and test losses for a misspecified linear regression with verbatim duplicates (\S\ref{sec:theory}) and recover the same non-monotonic peak in simulation, giving a simple statistical analogue for the empirical pattern observed in the language-model sweeps under exact document replay at fixed repeated-token fraction.

\section{Related Work}
\label{sec:related-work}

We position our work against three related lines of research; Appendix~\ref{app:related-work} extends the discussion.

\textbf{Compute-optimal scaling and the Chinchilla scaling law.} \citet{kaplan2020scaling} established the power-law form for transformer loss as a function of compute. \citet{hoffmann2022training} corrected the optimal $(N,T)$ allocation, where $N$ is the parameter count and $T$ is the token count. \citet{porian2024resolving,besiroglu2024chinchilla} examine the robustness of these fits, \citet{sardana2024beyond} extend them to inference-aware budgets, and \citet{gadre2024language} show reliable extrapolation through aggressive over-training. We use this functional form as our fitted no-repetition reference curve. 

\textbf{Deduplication, memorization, and statistical accounts of overfitting.} A large literature studies deduplication and memorization for privacy and contamination reasons~\citep{lee2022deduplicating,kandpal2022deduplicating,carlini2023quantifying,abbas2023semdedup,tirumala2023d4,deng2024unveiling,schaeffer2026quantifying}. Our setting is distinct in two ways. First, we exclude the evaluation split from training and from all repeated pools by the train/test split, so any harm we observe must come from a distorted effective training distribution. Second, we hold the fraction of repeated tokens fixed and vary only their concentration, isolating the effect of repetition structure. Our theoretical analysis connects to classical results on double descent and benign overfitting~\citep{belkin2019reconciling,hastie2022surprises,bartlett2020benign,nakkiran2020deep}, specialized to literal sample duplication. We are not aware of prior work using this block-covariance view to analyze literal duplication.


\textbf{Repeated data in language model pretraining.} \citet{hernandez2022scaling} repeat a small fraction of training data and observed a non-monotonic test-loss curve. They framed the damage as a reduction in effective parameter count and connected it to degradation of induction heads. \citet{muennighoff2023scaling} study the complementary regime in which the entire corpus is uniformly repeated, and find that up to roughly four epochs are nearly free. \citet{xue2023repeat,maini2024rephrasing,komatsuzaki2019one} examine related repetition, rephrasing, and epoch strategies. Recent work further argues that semantic duplication is itself scale-dependent~\citep{kazdan2026scale} and proposes explicit overfitting penalties for data-constrained scaling~\citep{lovelace2026prescriptive}. We sit between these two regimes, and extend \citet{hernandez2022scaling} by measuring damage as $\CEG$ and $\CEL$ rather than as a reduction in effective parameter count: for each repeated-data run, we ask how much compute a no-repetition run would need to reach the same loss. We also run the sweep under Chinchilla-style token and compute budgets while holding the repeated-token fraction fixed at $f=0.1$ and varying its concentration through $\Rep$.


\section{Methods}
\label{sec:methods}

Repeated data is increasingly hard to avoid during pretraining.  Thus, it becomes essential to understand which configurations of repeated data are dangerous and how much compute they waste. We fix the repeated-token fraction--or equivalently, repeated-token compute fraction--at $f=0.1$ and vary only its structure. By holding the fraction of tokens dedicated to repeats constant, we isolate the impact of repetition \emph{structure} from that of repetition \emph{amount}, letting us identify the most harmful configurations at fixed compute.                     

\paragraph{Setup.} We train Qwen3-style decoder-only transformers~\citep{yang2025qwen3,vaswani2017attention,su2024roformer} on FineWeb-Edu-Dedup~\citep{penedo2024fineweb}, using six parameter counts $N \in \{34, 48, 63, 93, 153, 344\}\mathrm{M}$. We define this architecture in more detail in Appendix~\ref{app:architecture}.  Each run is parameterized by a model size $N$ and an overtraining multiplier $\OT \in \{0.25, 0.5, 1, 2, 4\}$ with larger models run on a subset of this grid due to compute constraints. Setting $\OT=1$ corresponds to 20 tokens per parameter, following Chinchilla-style scaling~\citep{hoffmann2022training}, while $\OT>1$ implies overtraining relative to the compute-optimal token amount. For a given $(N,\OT)$ pair, the total number of training tokens is $T = 20 \cdot \OT \cdot N$. Using the standard dense-transformer estimate of $6N$ FLOPs per token~\citep{kaplan2020scaling,hoffmann2022training,sardana2024beyond,porian2024resolving,gadre2024language}, the total training compute is
\begin{equation}                                                                         
      C = 6NT = 120 \cdot \OT \cdot N^2.    
      \label{eq:budget}  
  \end{equation}        
Thus, within each $(N,\OT)$ sweep, both the token budget $T$ and compute budget $C$ are fixed.                                                                                           
\paragraph{Repeated-pool construction.}  
  For each repeated-data run, $(1-f)T$ tokens are drawn from non-repeated documents.            
  Let $\Dr$ denote the number of unique tokens in the repeated pool, and let $\Rep$ denote the number of times each repeated document is replayed.                                         
  The repeated-token budget is therefore      
  \[
  fT \approx \Rep \Dr
  \]
  so
  \begin{equation}
      \Dr \approx \frac{fT}{\Rep}
          = \frac{2 \cdot \OT \cdot N}{\Rep}.
      \label{eq:dr}
  \end{equation}
  Increasing $\Rep$ does not change the amount of repeated material we train on; it concentrates the same 10\% repeated-token budget onto a smaller pool.
  The $\Rep$ view answers how many times repeated documents are replayed, while the $\Dr$ view answers how large the repeated corpus is.
  Both are needed because repetition damage depends on their interaction.

  The repeated documents are sampled at document granularity and are disjoint from the non-repeated training documents. Copies of repeated documents are randomly interleaved with the non-repeated stream during training. This setup gives us a controlled setting to study repetition---which is unavoidable in any pretraining setup---and identify the configurations that cause the most damage. See Appendix~\ref{app:sampling} for more details on the sampling protocol.

\paragraph{Evaluation and baselines.}
We evaluate every model on a fixed held-out split of approximately 150M tokens, constructed with a fixed train/test seed. This evaluation split is excluded from all training data and from all repeated pools by the fixed train/test split. For each completed $(N,\OT)$ sweep, we also train a no-repetition baseline. These baselines serve two purposes. First, the baseline at the same $(N,\OT)$ gives the per-sweep reference for measuring the fractional eval-loss increase caused by repetition. Second, the six $\OT=1$ no-repetition baselines calibrate the no-repetition Chinchilla scaling law $L(C)=E+KC^{-\gamma}$ used in \S\ref{sec:finding-3} to convert eval loss into $\CEG$ and $\CEL$. Additional implementation and evaluation details are given in Appendix~\ref{app:training-details}.

\section{Results}


\begin{figure}[t]
  \centering
  \includegraphics[width=\linewidth]{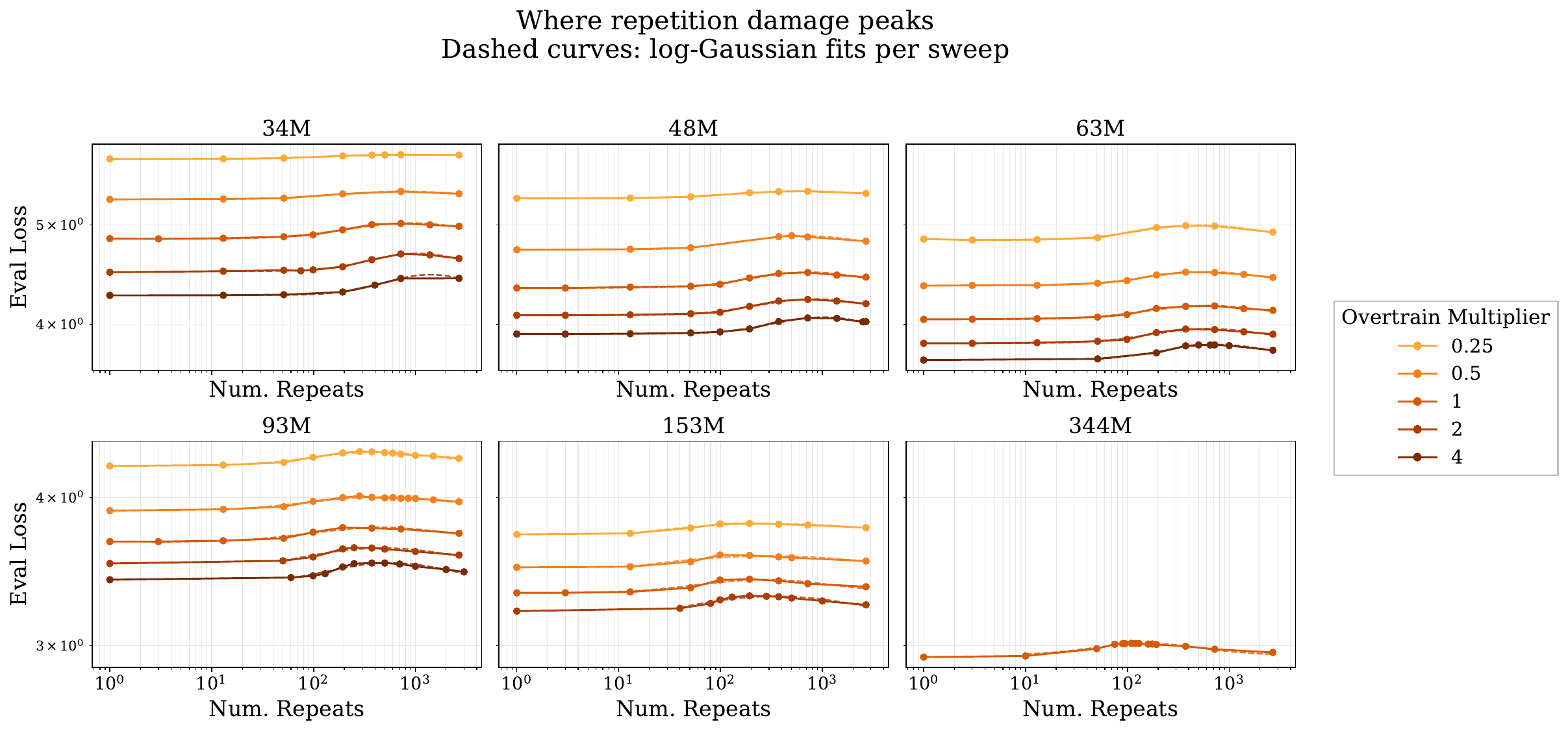}
  \caption{
    Gaussian fits to eval loss as a function of repeat count.
    Each panel fixes a model size $N$, and each curve corresponds to an overtraining multiplier $\OT$.
    The fitted peak gives $\Rep^{\mathrm{peak}}$, the repeat count at which eval loss is largest for that $(N, \OT)$ sweep.
    Across all model sizes, eval loss is maximized at an intermediate repeat count.
  }
  \label{fig:gaussian-fits-loss-r}
\end{figure}

We characterize how eval loss depends on the structure of the repeated subset (\S\ref{sec:finding-1}), extract a model-size scaling law for the worst-case configuration (\S\ref{sec:finding-2}), translate the resulting loss differences into  $\CEG$ and $\CEL$ via the no-repetition Chinchilla scaling law (\S\ref{sec:finding-3}), and explain the same non-monotonicity in a closed-form linear-regression model (\S\ref{sec:theory}--\S\ref{sec:linear-simulations}).

\subsection{Eval loss is non-monotonic in the repeat count}
\label{sec:finding-1}

For each $(N, \OT)$ pair we hold the repeated fraction $f = 0.1$ and the total compute $C$ fixed, varying only $\Rep$ along an iso-FLOP curve. Figure~\ref{fig:gaussian-fits-loss-r} plots eval loss against $\Rep$. The qualitative pattern is an intermediate-repeat peak with a sharpness that varies between sweeps. Across the completed sweeps, the raw maximum is $1.0$ to $4.2\%$ above the corresponding no-repetition baseline, with median $3.1\%$.  Measured relative to the larger of the two endpoints, no repeats and the largest measured $\Rep$, the peak prominence ranges from $0.7$ to $2.7\%$, with median $1.8\%$. The peaks are often broad. We therefore use the log-Gaussian fits in \S\ref{sec:finding-2} to estimate peak locations, rather than treating the discrete argmax as exact, because $\Rep$ is sampled on a finite, approximately logarithmic grid and the true maximum may fall between measured repeat counts.                               
        
\citet{hernandez2022scaling} reported a similar non-monotonic dependence in a fixed 100B-token-budget setting. We confirm that the same qualitative intermediate-repeat regime appears   
  under Chinchilla-style budgets, and quantify it using $\CEG$ and $\CEL$ in \S\ref{sec:finding-3}. The implication is that the most damaging repetition structure usually lies away from both extremes of the $\Rep$ range. Configurations with many repeats of a tiny pool, or few repeats
  of a large pool, generally produce smaller loss increases than configurations in between, though the recovery at large $\Rep$ is flatter in some sweeps. \S\ref{sec:theory} offers a statistical mechanism for why such a peak can arise.

\begin{figure}[t]
  \centering
  \includegraphics[width=0.9\linewidth]{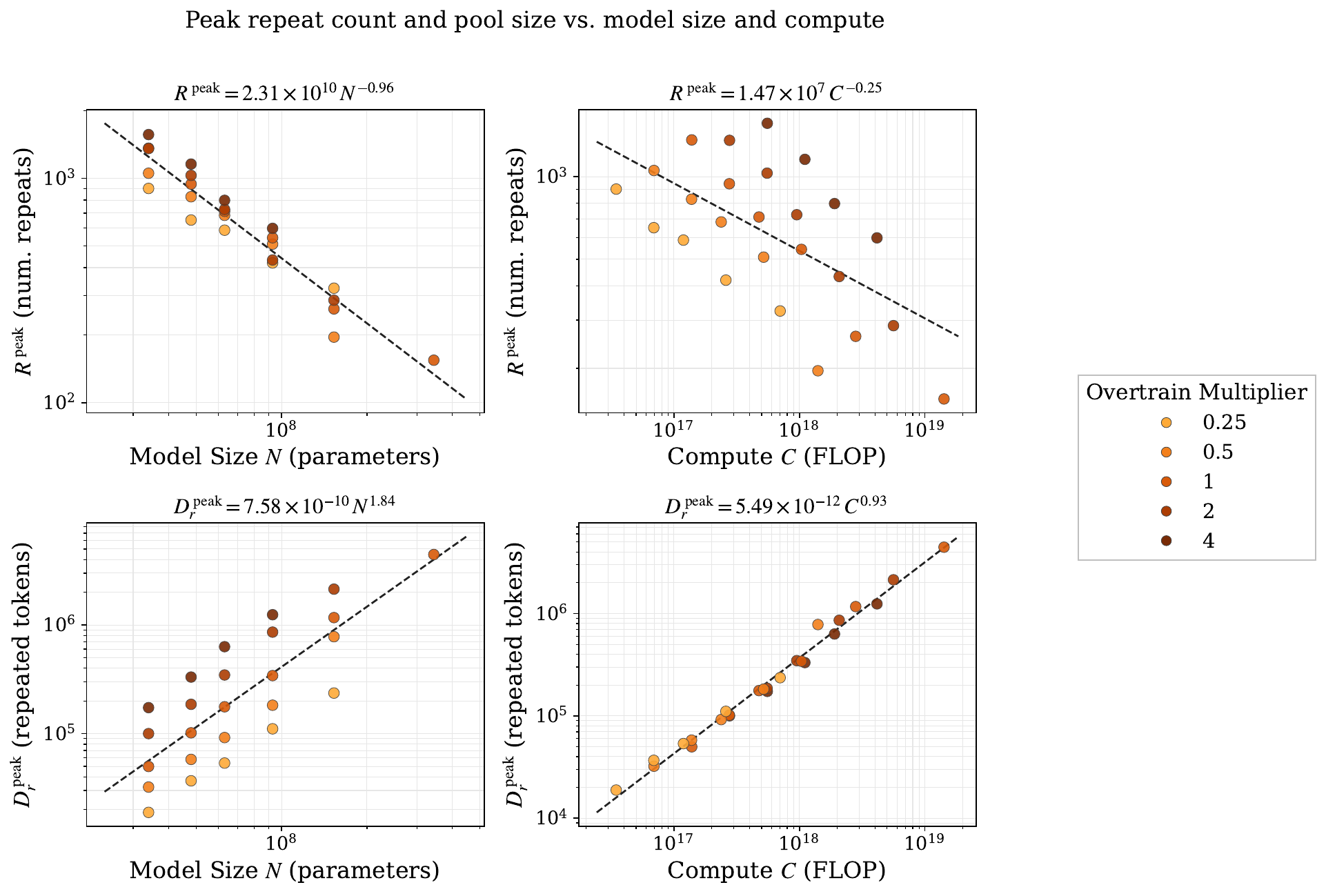}
  \caption{
    Scaling laws for the peak-damage regime.
    Top row, the repeat count at peak eval loss $\Rep^{\mathrm{peak}}$ decreases with model size (left) and compute (right).
    Bottom row, the repeated-pool size at peak eval loss $\Dr^{\mathrm{peak}}$ increases with both.
    These fits predict the most damaging repetition structure for a given training budget.
  }
  \label{fig:peak-scaling-laws}
\end{figure}

  \subsection{Peak locations are consistent with a power-law trend in model size}
  \label{sec:finding-2}

  \S\ref{sec:finding-1} established that loss is maximized at some intermediate $\Rep$. To summarize how this peak shifts with scale, we fit a simple empirical relationship between the
  estimated peak location and model size. We extract continuous peak locations by fitting a log-Gaussian to eval loss as a function of $\log_{10} \Rep$ for each $(N, \OT)$ pair,
  \begin{equation}
    L(x) = b + A \exp\!\left( -\frac{(x - \mu)^2}{2\sigma^2} \right),
    \qquad \Rep^{\mathrm{peak}} = 10^{\mu}.
    \label{eq:gaussian-fit}
  \end{equation}

This fit captures the single peak we observe in log-repeat space. We then fit power laws to the estimated $\Rep^{\mathrm{peak}}$ values across the completed $(N,\OT)$ grid and convert them to repeated-pool sizes using \eqref{eq:dr}, giving:
  
  \begin{align}
    \Rep^{\mathrm{peak}} &= 2.31 \times 10^{10} N^{-0.96},
    & \Dr^{\mathrm{peak}} &= 7.58 \times 10^{-10} N^{1.84}, \label{eq:peak-N}\\
    \Rep^{\mathrm{peak}} &=  1.47 \times 10^{7} C^{-0.25},
    & \Dr^{\mathrm{peak}} &= 5.49 \times 10^{-12} C^{0.93}. \label{eq:peak-C}
  \end{align}
  Figure~\ref{fig:peak-scaling-laws} plots these four related views. We note two caveats to this scaling law.  Because we fix the repetition token budget at $10\%$ and the number of repeated tokens at the peak grows more quickly than the compute budget, necessarily the peak will eventually (for large models) cross $0.1$T, which implies a number of repeats below $1$ under our setup.  Thus, for very large models, we interpret the scaling law as a statement that the model's memorization capacity grows faster than compute rather than an exact predictive tool.  

  \paragraph{Observed trend.}
  Within the measured range, larger models tend to reach their largest loss increase at fewer repeats of a larger repeated pool. The Qwen3-style 34M-parameter model peaks at $\Rep \approx 1400$ with $\Dr
  \approx 5 \times 10^4$ tokens, while the Qwen3-style 344M-parameter model peaks at $\Rep \approx 155$ with $\Dr \approx 4.5 \times 10^6$ tokens \cite{yang2025qwen3}. We also do not observe a strong
  dependence on training duration over the completed grid: the $\OT \in \{0.25, 0.5, 1, 2, 4\}$ sweeps largely fall near the same trend in $N$. Thus, the fitted trend provides a useful heuristic for identifying repetition regimes that may be especially harmful within the studied scale range.


\subsection{Repetition can cause an
\texorpdfstring{$\mathcal{O}(1)$}{O(1)}
Compute-Equivalent Loss}
\label{sec:finding-3}

The 2 to 4\% loss increases reported in \S\ref{sec:finding-1} and \S\ref{sec:finding-2} translate into different compute gaps depending on where the run sits on the no-repetition Chinchilla curve. \citet{hernandez2022scaling} addressed this by reporting damage as a reduction in effective parameter count, but that framing predates Chinchilla-style scaling and is hard to compare across model sizes and training durations. We replace it with a compute-equivalent ratio. We ask how much compute a no-repetition run would need to match the loss of a repeated-data run.
\begin{figure}[t!] 
\centering
\includegraphics[width=0.4\linewidth]{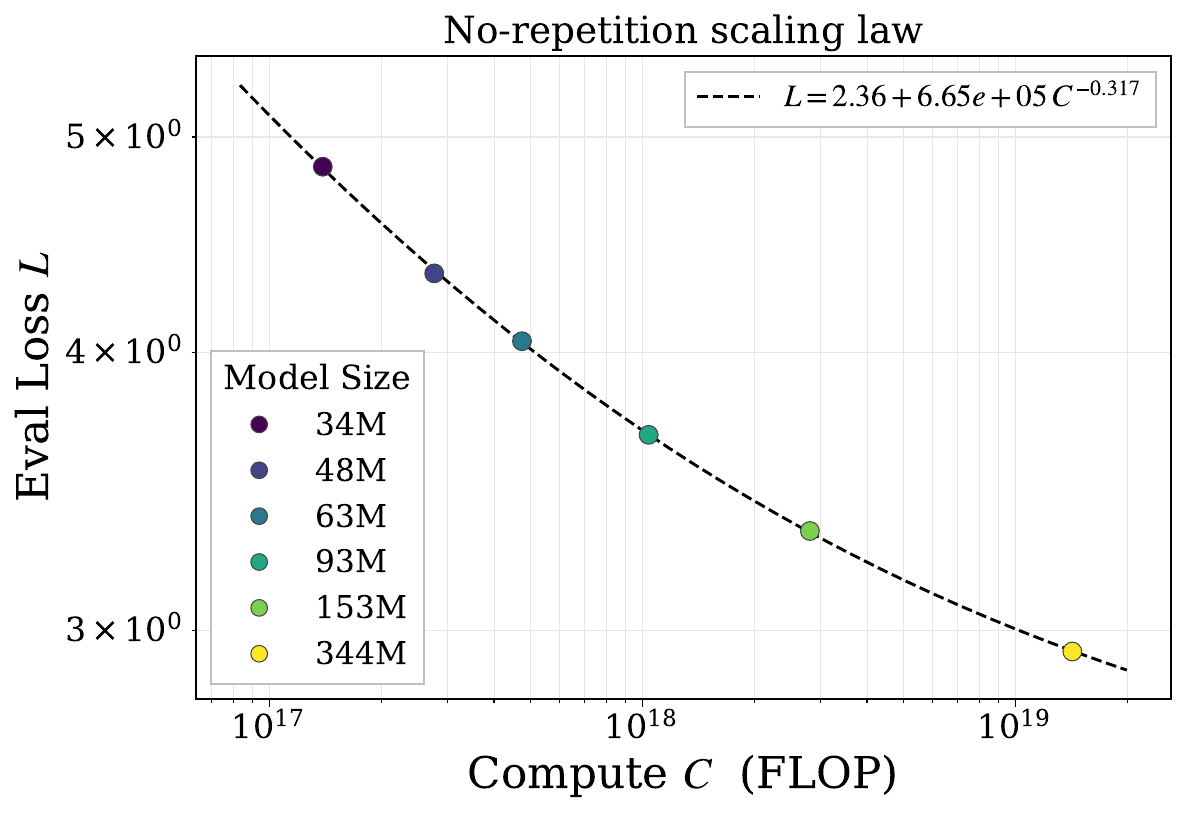}
\caption{
  No-repetition scaling law fit.
  We fit $L(C) = E + KC^{-\gamma}$ to the six OT=1, $\Rep = 1$ baselines.
  This scaling law converts any eval loss into the equivalent no-repetition compute, enabling the $\CEG$ and $\CEL$ metrics.
}
\label{fig:compute-savings-frontier}
\end{figure}

We fit the no-repetition Chinchilla scaling law $L(\flops) = E + K \flops^{-\gamma}$ to the six $\OT{=}1$, $\Rep{=}1$ baselines~\citep{hoffmann2022training,gadre2024language,porian2024resolving}, obtaining
\begin{equation}
  L(\flops) = 2.365 + 6.647 \times 10^5 C^{-0.317} .
  \label{eq:frontier-fit}
\end{equation}

The fit is shown in Figure~\ref{fig:compute-savings-frontier}. Here $E$ is the irreducible-loss floor that the model approaches in the limit of infinite compute, $K$ sets the overall vertical scale, and $\gamma > 0$ is the rate at which loss decreases with compute. Inverting the law gives $\flops^\star(L) = (K/(L - E))^{1/\gamma}$, the amount of compute a no-repetition run on the law would need to reach a measured loss $L$. The \emph{Compute-Equivalent Gain} of a repeated-data run is
\begin{equation}
  \CEG = \frac{\flops^\star(L)}{\flops_{\mathrm{actual}}},
  \qquad
  \CEL = 1-\CEG .
  \label{eq:ceg-cel}
\end{equation}

$\CEG=1$ matches the no-repetition law. When $\CEG<1$, the run reaches the loss of a no-repetition run trained with only $\CEG \cdot \flops_{\mathrm{actual}}$ FLOPs, and $\CEL=1-\CEG$ is the Compute-Equivalent Loss.
  
\begin{figure}[htbp!]
\centering
\includegraphics[width=\linewidth]{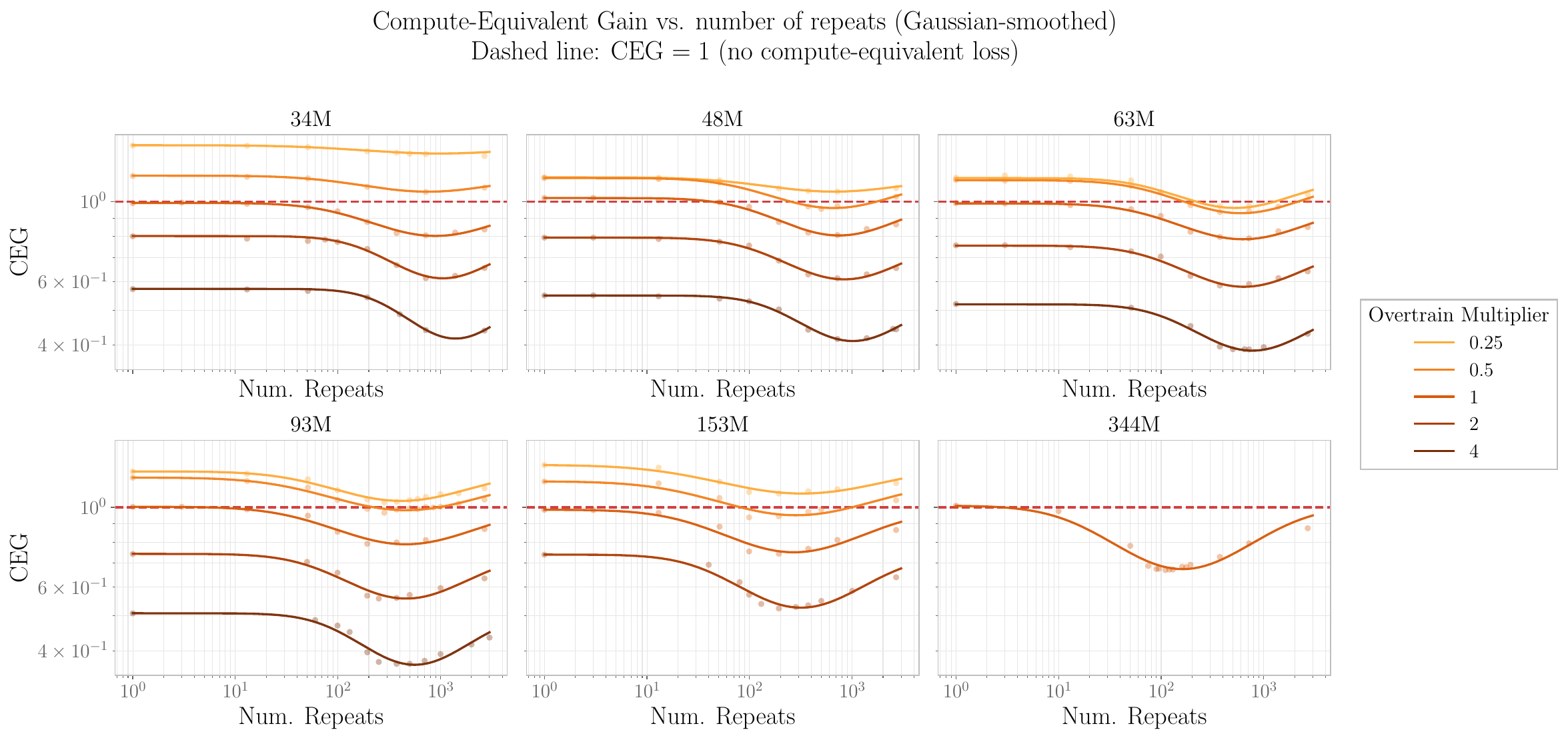}
\caption{
  Compute-Equivalent Gain as a function of repeat count, by model size and overtraining multiplier.
  $\CEG = 1$ (dashed line) matches the no-repetition reference.
  $\CEG$ falls at intermediate repeat counts and partially recovers at large ones, revealing a worst-case repetition structure for each sweep.
  Damage grows with model size. The Qwen3-style 344M-parameter model reaches $\CEG\approx0.67$, corresponding to $\CEL\approx0.33$.
}
\label{fig:compute-savings-repeats}
\end{figure}

Figure~\ref{fig:compute-savings-repeats} shows $\CEG$ as a function of $\Rep$. At $\OT{=}1$, the worst repeat settings increase $\CEL$ to $0.19, 0.19, 0.21, 0.21, 0.26, 0.33$ for $N \in \{34, 48, 63, 93, 153, 344\}$M respectively. \textbf{At the largest scale, peak repetition produces $\bm{\CEL\approx0.33}$.} Two patterns are notable. First, the 2 to 4\% loss bump translates into $\CEL$ values of roughly $0.19$ to $0.33$ because the no-repetition scaling law is shallow. A small loss gap maps to a large compute gap, and the loss-space view systematically understates the practical cost. Second, varying $\OT$ shifts the level of $\CEG$ but leaves the location of the peak in $\Rep$ approximately unchanged. The worst-case repetition structure persists across training durations.

\paragraph{Sensitivity to the loss floor.}
Equation~\eqref{eq:ceg-cel} contains the factor $(L - E)^{-1/\gamma}$, which diverges as $L$ approaches the fitted floor $E$. With $\gamma \approx 0.32$, a $1\%$ shift in $(L-E)$ produces a $\approx 3\%$ relative shift in $\CEG$. The 344M peak run sits at $L - E \approx 0.65$ nats, so the headline 33\% number is robust to leave-one-out perturbations of the scaling-law fit. Smaller models live closer to $E$ and inherit larger uncertainty. The scaling law is fit on six points, at the boundary of identifiability for the three-parameter Chinchilla form~\citep{besiroglu2024chinchilla,porian2024resolving}. We therefore treat the absolute $\CEG$ and $\CEL$ values as point estimates and the qualitative non-monotonicity as the robust finding.

\subsection{A statistical model of repetition damage}
\label{sec:theory}

The non-monotonic peak in eval loss could plausibly be a transformer-specific artifact, a memorization quirk arising from attention, depth, or optimization dynamics. We suggest that it may reflect a more general statistical effect. A finite-capacity learner trained on duplicated samples from a richer distribution can trade fit on the repeated set against generalization beyond the repeated samples, and this tradeoff can produce non-monotonic behavior in the duplication count. We illustrate this possibility in misspecified linear regression, deriving closed-form conditional risks and recovering the same qualitative peak in simulations.

\paragraph{Setup.} The data-generating process is a high-dimensional linear model with isotropic Gaussian inputs. Inputs are $x \sim \mathcal{N}(0, I_p)$ in $\mathbb{R}^p$, with noiseless labels $y = x^\top \beta$ and a fixed coefficient vector $\beta \in \mathbb{R}^p$. The learner observes only the first $m < p$ coordinates of $x$, so the model is misspecified. Decompose $x = (x_{\mathrm{in}}, x_{\mathrm{out}})$ and $\beta = (\beta_{\mathrm{in}}, \beta_{\mathrm{out}})$, with $x_{\mathrm{in}}, \beta_{\mathrm{in}} \in \mathbb{R}^m$ observed and $x_{\mathrm{out}}, \beta_{\mathrm{out}} \in \mathbb{R}^{p-m}$ unobserved. The same isotropic distribution governs the test point. The unobserved coordinates carry real predictive signal, so finite-sample correlations between $x_{\mathrm{in}}$ and $x_{\mathrm{out}}$ can be mistaken for useful structure. The training set contains $n$ unique examples plus a block of $d$ examples each repeated $r$ times, for a total of $n + rd$ rows.

\paragraph{Block-diagonal noise covariance.} Let $X_{u,\mathrm{in}}$ and $X_{d,\mathrm{in}}$ stack the observed features for the unique and repeated examples respectively, and let $C_u = X_{u,\mathrm{in}}^\top X_{u,\mathrm{in}}$ and $C_d = X_{d,\mathrm{in}}^\top X_{d,\mathrm{in}}$ be the corresponding observed-feature Gram matrices. The expanded observed-feature Gram of the full training set is $B_r = C_u + r C_d$. The restricted OLS estimator decomposes as $\hat\beta_{\mathrm{in}} = \beta_{\mathrm{in}} + a_r$, where $a_r$ is an aliasing term that arises from fitting the unobserved signal through the observed coordinates. The crucial step is that the $r$ copies of each repeated example share a single unobserved-feature realization, so the conditional covariance of $z = X_{\mathrm{out}} \beta_{\mathrm{out}}$ given $X_{\mathrm{in}}$ is the block-diagonal direct sum
\begin{equation}
    \Sigma_r \;=\; I_n \;\oplus\; \bigoplus_{i=1}^{d} \mathbf{1}_r \mathbf{1}_r^\top \;\in\; \mathbb{R}^{(n+rd) \times (n+rd)},
    \label{eq:Sigma-r}
\end{equation}
where $\mathbf{1}_r \in \mathbb{R}^r$ is the all-ones vector and the symbol $\oplus$ denotes block-diagonal direct sum. The unique block is the $n \times n$ identity, reflecting independent unobserved features across unique examples. Each repeated block is the rank-one $r \times r$ matrix $\mathbf{1}_r \mathbf{1}_r^\top$, reflecting that the $r$ copies of document $i$ share a single $x_{\mathrm{out},i}$. Distinct repeated documents are uncorrelated. Multiplying through gives $X_{\mathrm{in}}^\top \Sigma_r X_{\mathrm{in}} = C_u + r^2 C_d$. The extra factor of $r$ is what makes duplication qualitatively different from adding more independent samples. Conditioning on $X_{\mathrm{in}}$, the expected train and test errors are
\begin{align}
    \mathbb{E}[L_{\mathrm{train}} \mid X_{\mathrm{in}}] &= \frac{\|\beta_{\mathrm{out}}\|_2^2}{n+rd} \left[n+rd - \operatorname{tr} \!\left( (C_u+r^2C_d)(C_u+rC_d)^{-1} \right)\right], \label{eq:train-loss}\\
    \mathbb{E}[L_{\mathrm{test}} \mid X_{\mathrm{in}}] &= \|\beta_{\mathrm{out}}\|_2^2 \left[1 + \operatorname{tr}\!\left((C_u+rC_d)^{-1}(C_u+r^2C_d)(C_u+rC_d)^{-1}\right)\right].
    \label{eq:test-loss}
\end{align}

Equations~\eqref{eq:train-loss}--\eqref{eq:test-loss} are closed-form conditional risks. The non-monotonic behavior is a simulation-supported mechanism explored in \S\ref{sec:linear-simulations}. Appendix~\ref{app:linear-repetition-theory} gives the derivation. 

\paragraph{What the formulas suggest.} The analogy to the language-model setting is that $m$ plays the role of model capacity and $d$ corresponds to the unique repeated-data pool $\Dr$. In the regimes we simulate in \S\ref{sec:linear-simulations}, the test loss in \eqref{eq:test-loss} is non-monotonic in $r$ at fixed $(n, d, m)$, as follows. When $r$ is small, the repeated examples carry little extra weight and the predictor is only weakly affected. When $r$ is too large, the repeated block saturates the rank of $C_u + r^2 C_d$ relative to $C_u + r C_d$, and the test loss returns toward a memorize-and-isolate fixed point. The harmful middle regime appears when the repeated pool is both influential and too large to be harmlessly absorbed. The same formulas suggest that increasing $m$ shifts the peak to larger $d$ in the simulated setting, offering a statistical analogue consistent with the empirical signature in \eqref{eq:peak-N}. They also suggest weak dependence on the unique-sample count $n$ in this toy setting, consistent with the approximate OT-independence observed in \S\ref{sec:finding-2}.

\subsection{Simulations of the linear model}
\label{sec:linear-simulations}

\begin{figure}[htbp!]
    \centering
    \includegraphics[width=\linewidth]{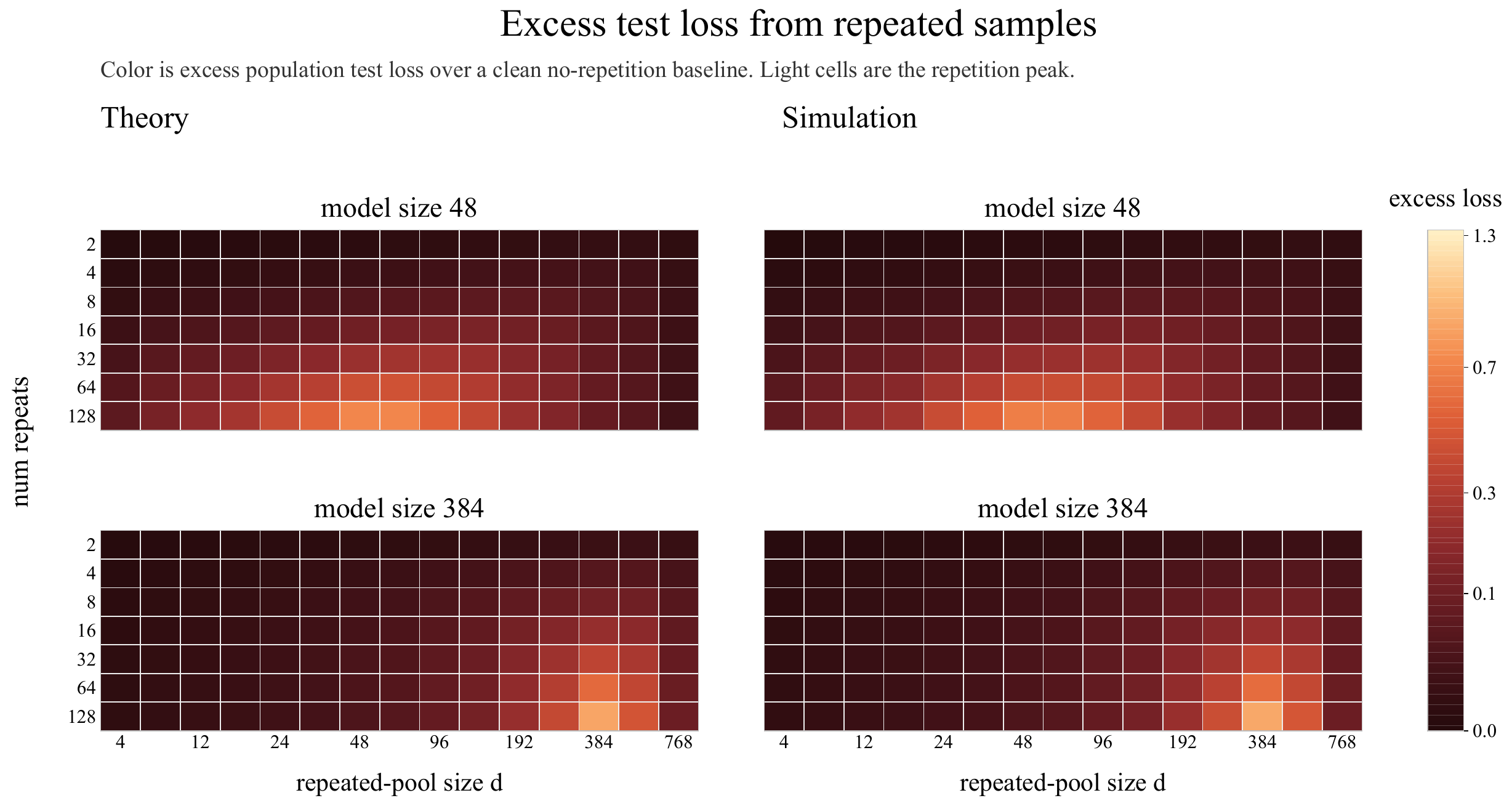}
    \caption{Excess test loss in misspecified linear regression with repeated samples. Rows vary the observed dimension $m$, and columns compare closed-form theory with direct OLS simulations. Within each panel, the $x$-axis is the repeated-pool size $d$, the $y$-axis is the repeat count $r$, and color indicates excess population test loss over a no-repetition baseline at the same $m$. The peak shifts to larger $d$ as $m$ increases, providing a statistical analogue consistent with the $\Dr^{\mathrm{peak}}(N)$ trend in Figure~\ref{fig:peak-scaling-laws}.
    }
    \label{fig:linear-loss-surfaces}
\end{figure}

We validate the closed-form theory with direct simulations. Inputs are sampled as $x \sim \mathcal{N}(0, I_p)$ and labels are generated by $y = x^\top \beta$ with $\|\beta\|_2 = 1$. The learner fits OLS on the first $m$ coordinates of $n$ unique samples plus $d$ samples each repeated $r$ times. For each $(m, d, r)$ we evaluate both \eqref{eq:test-loss} and the population test loss of the corresponding OLS solve. We report excess test loss over a no-repetition baseline at the same $m$. Figure~\ref{fig:linear-loss-surfaces} shows that the closed-form and simulation agree to within numerical precision, that excess loss is non-monotonic in the repeated-pool size $d$ at fixed $m$ and $r$, and that the peak shifts to larger $d$ as $m$ grows. This trend is consistent with the empirical $\Dr^{\mathrm{peak}}(N)$ relationship in \eqref{eq:peak-N}.

\begin{figure}[t]
    \centering
    \includegraphics[width=0.7\linewidth]{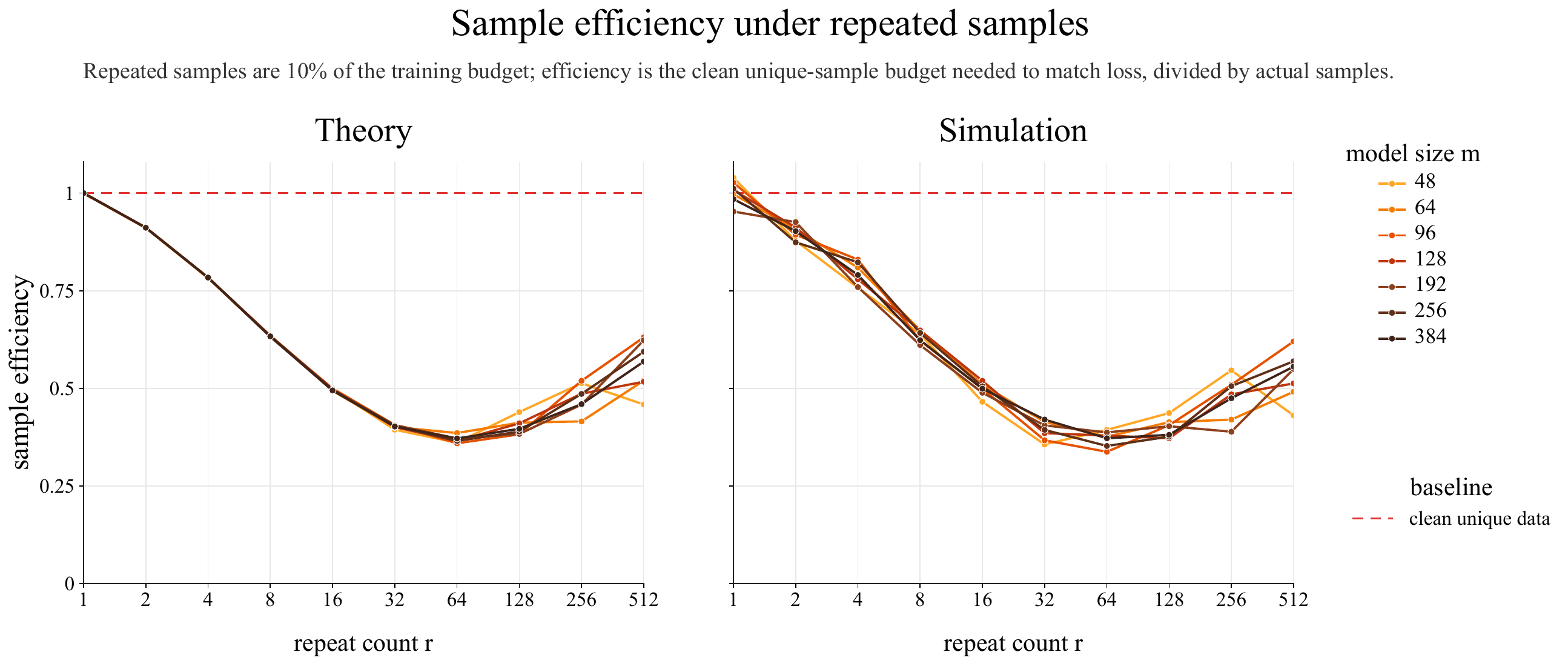}
    \caption{
    Sample efficiency under repeated samples. The repeated block accounts for $10\%$ of the training budget. For each repeat count $r$, $\mathrm{SE}$ is the unique-sample budget needed to match the repeated-data test loss, divided by the actual sample budget. Theory and simulation both show non-monotonic efficiency loss, mirroring Figure~\ref{fig:compute-savings-repeats}.
    }
    \label{fig:linear-sample-efficiency}
\end{figure}

We also compute a sample-efficiency analogue of $\CEG$. For each repeated-data run, we estimate the no-repetition unique sample budget $N^\star_{\mathrm{clean}}$ required to match its test loss, and define $\mathrm{SE} = N^\star_{\mathrm{clean}} / N_{\mathrm{actual}}$, with the repeated block again accounting for $10\%$ of the training budget. Figure~\ref{fig:linear-sample-efficiency} shows that $\mathrm{SE}$ falls sharply at intermediate $r$ and partially recovers at extreme $r$, mirroring the $\CEG$ curve in Figure~\ref{fig:compute-savings-repeats} for language models. Figure~\ref{fig:peak-heat} shows where the worst repeated-pool size occurs for each model size and repeat count. The closed-form linear model thus captures the same qualitative phenomenon, suggesting that the peak is a generic statistical feature of repeated samples in misspecified models.

\section{Conclusion}

Pretraining is now data-constrained, and this increases risk of residual repetition. We studied exact document-level repetition in a controlled setting, holding the repeated-token fraction fixed at $f=0.1$ and varying only the repeat count $\Rep$. At fixed compute, eval loss is worst at an intermediate repeat count: a moderately sized repeated pool replayed many times can hurt more than either a tiny pool replayed many times or a larger pool replayed only a few times. The estimated peak locations follow a clear trend over the model sizes we test: larger models tend to peak at fewer repeats of larger repeated pools. We treat this as an empirical summary of our sweep. Under our fitted no-repetition scaling law, the most damaging repeat setting at our largest scale had $\CEG\approx0.67$, corresponding to $\CEL\approx0.33$, meaning it reaches the loss a no-repetition run would reach with about two-thirds of the compute. The linear-regression model gives a simple analogue for this pattern. Our results show that it is not enough to report how much data is duplicated: at the same duplicated-token fraction $f$, the number of times each duplicate appears can substantially change the compute cost.


\newpage

\medskip






\bibliographystyle{unsrtnat}
\bibliography{bib}

\appendix

\section{Comprehensive related work}
\label{app:related-work}

We expand here the discussion sketched in \S\ref{sec:related-work}, organized along five threads.

\paragraph{Repeated data in language model pretraining.}
Our closest predecessor is \citet{hernandez2022scaling}, who train transformers with a small fraction of repeated data and observe a non-monotonic test-loss curve. They frame the damage as a reduction in effective parameter count and connect it to mechanistic changes in induction heads~\citep{biderman2023emergent,allenzhu2024physics}. The complementary regime, in which the entire corpus is uniformly repeated, is studied by \citet{muennighoff2023scaling}. They find that up to roughly four epochs are nearly as useful as fresh data and that an additive overfitting term in the Chinchilla loss captures the rest. \citet{xue2023repeat} reach a similar four-epoch threshold from a different angle. Other work in this family includes \citet{komatsuzaki2019one}, who studies one-pass-vs-multi-pass tradeoffs at smaller scale, and \citet{maini2024rephrasing}, who replaces literal repeats with paraphrases generated by a teacher model. \citet{tirumala2022memorization} and \citet{lesci2024causal} study how memorization grows with the number of times an example is seen during training. \citet{kazdan2026scale} argue that semantic duplication is itself scale-dependent, in the sense that larger models recognize more documents as duplicates. \citet{lovelace2026prescriptive} fit a one-parameter overfitting penalty within Chinchilla scaling that is closely related in spirit to our $\CEG/\CEL$ metric. Our setting differs from all of these in three ways. First, we hold a fixed minority fraction $f = 0.1$ of training tokens repeated, mimicking the residue of imperfect deduplication and complementing the all-or-nothing regimes of \citet{muennighoff2023scaling}. Second, we measure damage in compute-equivalent units derived from a fitted Chinchilla scaling law, sharpening the effective-parameter view of \citet{hernandez2022scaling} into a quantity practitioners directly allocate. Third, we extract a closed-form scaling law for the worst-case configuration as a function of model size, which earlier work has left open.

\paragraph{Compute-optimal scaling.}
The power-law form for transformer loss was established by \citet{kaplan2020scaling} and refined by \citet{hoffmann2022training}, whose Chinchilla fit is the basis for our no-repetition reference curve. Subsequent work has tested the robustness and generalizability of the Chinchilla functional form. \citet{besiroglu2024chinchilla} reanalyze the original Chinchilla fits, \citet{porian2024resolving} reconcile competing exponents across studies, \citet{sardana2024beyond} extend the framework to inference-aware budgets, and \citet{gadre2024language} show that Chinchilla-style fits extrapolate reliably under aggressive over-training in the regime we use. \citet{caballero2023broken,bahri2024explaining,bordelon2024dynamical,henighan2020scalingauto} provide alternative parametric forms and theoretical accounts. \citet{tay2022scaling} document the model-family dependence of fitted exponents, supporting our caveat in the Limitations that exponents may shift across architecture families.

\paragraph{Deduplication, memorization, and benchmark contamination.}
Deduplication has been a central tool of pretraining-corpus construction since at least \citet{lee2022deduplicating}, who show that exact and near-duplicate removal improves training efficiency and reduces verbatim regurgitation. \citet{kandpal2022deduplicating} document a superlinear amplification of privacy risk by duplicates, and \citet{carlini2023quantifying,carlini2021extracting} relate memorization to model scale and duplicate count. Semantic deduplication was advanced by \citet{abbas2023semdedup} and \citet{tirumala2023d4}, who use embedding clusters to remove near-duplicate documents that elude exact-hashing pipelines. Aggregate surveys include \citet{deng2024unveiling} and the data-selection survey of \citet{albalak2024survey}. A related but distinct concern is benchmark contamination, where evaluation data leaks into training~\citep{deng2024unveiling,schaeffer2026quantifying,oren2024proving,magar2022data,jacovi2023stop}. \citet{schaeffer2026quantifying} show that even a single test-set replica can drive measured loss below the no-contamination irreducible floor. We deliberately exclude evaluation documents from training and repeated pools by the fixed train/test split, so the harm we observe comes from a distorted effective training distribution.

\paragraph{Pretraining corpora and data selection.}
Our experiments use FineWeb-Edu-Dedup~\citep{penedo2024fineweb}. Closely related curated web corpora include DataComp-LM~\citep{li2024datacomp}, Dolma~\citep{soldaini2024dolma}, RedPajama-v2~\citep{weber2024redpajama}, and RefinedWeb~\citep{penedo2023refinedweb}. \citet{longpre2024pretrainers} survey the broader pretrainer's-data landscape. On the data-selection side, \citet{sorscher2022beyond} show that careful pruning can break power-law scaling, \citet{marion2023less} study perplexity-based pruning, \citet{xie2023doremi} optimize domain mixtures, and \citet{goyal2024scaling} extend scaling laws to data-quality interventions. Our results complement this view by studying a \emph{negative} form of data selection, asking which residual repetition structures to avoid. The two perspectives are quantitatively connected through the same Chinchilla scaling law, since both ultimately translate dataset choices into a position on a no-intervention scaling curve.

\paragraph{Statistical accounts of overfitting.}
The closed-form analysis in \S\ref{sec:theory} sits in the literature on benign overfitting and double descent~\citep{belkin2019reconciling,nakkiran2020deep,hastie2022surprises,bartlett2020benign,belkin2020two,advani2020highdimensional,tsigler2023benign,mei2022generalization}, which studies non-monotonic generalization in over-parameterized linear and kernel models. Most of this literature varies model dimension, sample size, or interpolation. Literal verbatim duplication of a subset of observations and the resulting block-diagonal noise covariance \eqref{eq:Sigma-r} are, to our knowledge, novel in this setting. Our derivation isolates the harm from duplication itself, controlling for the change in dataset size that would otherwise confound the effect.


\section{Architecture.}
\label{app:architecture}

  We instantiate all models from scratch using the Qwen3 decoder architecture family~\citep{yang2025qwen3}. Across model sizes, we vary depth, hidden width, and feed-forward width, while
  holding the remaining architectural settings fixed. All models use rotary position embeddings~\citep{su2024roformer}, RMSNorm, SwiGLU feed-forward layers, grouped-query attention,
  untied input/output embeddings, BF16 training, and FlashAttention-2. The training sequence length is 2048 tokens. The maximum position length is 32768, the vocabulary size is 151670,
  the attention head dimension is 128, and all models use 32 attention heads and 32 key-value heads. Table~\ref{tab:qwen-arch} reports the exact configurations and parameter counts. Non-
  embedding parameters exclude both the token embedding matrix and the untied output LM head; total parameters include both.

  \begin{table}[t]
  \caption{Qwen3-style model configurations used in the experiments.}
  \label{tab:qwen-arch}
  \centering
  \small
  \setlength{\tabcolsep}{3.5pt}
  \resizebox{\linewidth}{!}{
  \begin{tabular}{lrrrrrrrrrr}
  \toprule
  Model & Layers & $d_{\mathrm{model}}$ & $d_{\mathrm{ff}}$ & Heads & KV heads & $d_{\mathrm{head}}$ & Train ctx. & Vocab & Non-emb. params & Total params \\
  \midrule
  34M  & 3  & 96  & 256  & 32 & 32 & 128 & 2048 & 151670 & 4,941,216   & 34,061,856 \\
  48M  & 4  & 128 & 512  & 32 & 32 & 128 & 2048 & 151670 & 9,177,216   & 48,004,736 \\
  63M  & 5  & 160 & 512  & 32 & 32 & 128 & 2048 & 151670 & 14,339,040  & 62,873,440 \\
  93M  & 6  & 224 & 768  & 32 & 32 & 128 & 2048 & 151670 & 25,121,120  & 93,069,280 \\
  153M & 9  & 320 & 1024 & 32 & 32 & 128 & 2048 & 151670 & 56,041,664  & 153,110,464 \\
  344M & 14 & 576 & 1536 & 32 & 32 & 128 & 2048 & 151670 & 169,299,776 & 344,023,616 \\
  \bottomrule
  \end{tabular}
  }
  \end{table}

  \section{Repeated-pool construction details}
  \label{app:sampling}

  For each run, we first split FineWeb-Edu-Dedup into training and held-out evaluation documents using train/test split seed 0. The evaluation split is constructed before any repeated
  pools are selected, so evaluation documents are excluded from both the repeated and non-repeated training streams.

  Documents are tokenized with the Qwen3 \cite{yang2025qwen3} tokenizer, truncated to the training sequence length, and assigned EOS tokens before token counts are computed. Let $T$ be the target number of
  training tokens for a run, $f=0.1$ the target repeated-token fraction, and $\Rep$ the number of times each repeated document is replayed. The target unique repeated-pool size is
  \[
      D_{\mathrm{r}}^\star = \frac{fT}{\Rep},
  \]
  and the target non-repeated budget is $(1-f)T$.

  Documents are selected without replacement from the training split. We form a seeded random ordering of training documents using the run's shuffle seed. The repeated pool is the first
  prefix of this ordering whose cumulative token count reaches $D_{\mathrm{r}}^\star$. The non-repeated pool is then selected from the immediately following documents until its cumulative
  token count reaches $(1-f)T$. Thus selection is uniform over documents through a seeded shuffle, not token-weighted sampling, and the repeated documents are disjoint from the non-
  repeated documents.

  Because cutoffs occur at document boundaries, the realized unique repeated-pool size $\widehat{D}_{\mathrm{r}}$ and realized repeated-token fraction $\widehat{f}$ are approximate. If
  $L_{\max}$ is the maximum tokenized document length after truncation and EOS insertion, then
  \[
      D_{\mathrm{r}}^\star \le \widehat{D}_{\mathrm{r}} < D_{\mathrm{r}}^\star + L_{\max}.
  \]
  The non-repeated pool has the same one-document overshoot bound. In our preprocessing, $L_{\max}\le 2049$ because documents are truncated to length 2048 and an EOS token may be appended
  after truncation. Each document in the repeated pool is then inserted exactly $\Rep$ times, each non-repeated document is inserted once, and the resulting document-index list is
  shuffled before training.

  Throughout the paper, $\Dr=fT/\Rep$ denotes the target unique repeated-pool size. The approximation $fT \approx \Rep\Dr$ reflects this document-boundary rounding.

\section{Linear regression with repeated samples}
\label{app:linear-repetition-theory}

We derive the formulas used in Section~\ref{sec:theory}. Let the original training set contain $n$ unique examples and $d$ repeatable examples. The repeatable block is duplicated $r$ times, so the expanded training set has
\[
    N = n + rd
\]
rows. The learner observes only the first $m$ coordinates of each input. Write the expanded observed-feature matrix as $X_{\mathrm{in}}$ and the unobserved-feature matrix as $X_{\mathrm{out}}$. Labels are noiseless:
\[
    y = X_{\mathrm{in}}\beta_{\mathrm{in}} + X_{\mathrm{out}}\beta_{\mathrm{out}}.
\]

The restricted OLS estimator is
\[
    \hat{\beta}_{\mathrm{in}}
    =
    (X_{\mathrm{in}}^\top X_{\mathrm{in}})^{-1}X_{\mathrm{in}}^\top y
    =
    \beta_{\mathrm{in}} + a_r,
\]
where
\[
    a_r
    =
    (X_{\mathrm{in}}^\top X_{\mathrm{in}})^{-1}
    X_{\mathrm{in}}^\top X_{\mathrm{out}}\beta_{\mathrm{out}}.
\]
Thus $a_r$ is the aliasing term caused by fitting the unobserved part of the signal using the observed coordinates.

Let
\[
    C_u = X_{u,\mathrm{in}}^\top X_{u,\mathrm{in}},
    \qquad
    C_d = X_{d,\mathrm{in}}^\top X_{d,\mathrm{in}}.
\]
Then
\[
    X_{\mathrm{in}}^\top X_{\mathrm{in}} = C_u + rC_d .
\]

Now condition on $X_{\mathrm{in}}$ and take expectation over the unobserved features. Let
\[
    z = X_{\mathrm{out}}\beta_{\mathrm{out}}.
\]
Because repeated rows share the same unobserved coordinates, the covariance of $z$ over the expanded training set is not proportional to the identity. Instead,
\[
    \mathbb{E}[zz^\top \mid X_{\mathrm{in}}]
    =
    \|\beta_{\mathrm{out}}\|_2^2 \, \Sigma_r,
    \qquad
    \Sigma_r \;=\; I_n \;\oplus\; \bigoplus_{i=1}^{d} \mathbf{1}_r \mathbf{1}_r^\top ,
\]
where $\mathbf{1}_r \in \mathbb{R}^r$ is the all-ones vector and the $i$-th repeated block is the rank-one $r \times r$ matrix $\mathbf{1}_r \mathbf{1}_r^\top$. The unique block contributes the $n \times n$ identity (independent unobserved features); each repeated block has every entry equal to every other (the $r$ copies of document $i$ share a single $x_{\mathrm{out},i}$). The full matrix is $(n+rd) \times (n+rd)$ and block-diagonal; in particular, distinct repeated documents are uncorrelated. Therefore,
\[
    X_{\mathrm{in}}^\top \Sigma_r X_{\mathrm{in}}
    =
    C_u + r^2 C_d .
\]

For training loss, let
\[
    H =
    X_{\mathrm{in}}
    (X_{\mathrm{in}}^\top X_{\mathrm{in}})^{-1}
    X_{\mathrm{in}}^\top
\]
be the projection matrix onto the observed-feature span. The residual is $(I-H)z$, so
\[
    \mathbb{E}[L_{\mathrm{train}} \mid X_{\mathrm{in}}]
    =
    \frac{1}{N}
    \mathbb{E}[z^\top(I-H)z \mid X_{\mathrm{in}}].
\]
Using the covariance above,
\[
    \mathbb{E}[L_{\mathrm{train}} \mid X_{\mathrm{in}}]
    =
    \frac{\|\beta_{\mathrm{out}}\|_2^2}{N}
    \operatorname{tr}((I-H)\Sigma_r).
\]
Since $\operatorname{tr}(\Sigma_r)=N$ and
\[
    \operatorname{tr}(H\Sigma_r)
    =
    \operatorname{tr}
    \left(
        (C_u+r^2C_d)(C_u+rC_d)^{-1}
    \right),
\]
we obtain
\[
    \mathbb{E}[L_{\mathrm{train}} \mid X_{\mathrm{in}}]
    =
    \frac{\|\beta_{\mathrm{out}}\|_2^2}{N}
    \left[
        N
        -
        \operatorname{tr}
        \left(
            (C_u+r^2C_d)(C_u+rC_d)^{-1}
        \right)
    \right].
\]

For test loss, take a fresh example $(x_{\mathrm{in}}, x_{\mathrm{out}}) \sim \mathcal{N}(0, I_p)$ from the population, with $x_{\mathrm{in}}$ and $x_{\mathrm{out}}$ independent under the isotropic assumption. The prediction error is
\[
    x_{\mathrm{out}}^\top\beta_{\mathrm{out}}
    -
    x_{\mathrm{in}}^\top a_r .
\]
The two terms are independent and mean zero (the cross term vanishes by independence and zero mean of $x_{\mathrm{in}}$), so
\[
    \mathbb{E}[L_{\mathrm{test}} \mid X_{\mathrm{in}}]
    =
    \|\beta_{\mathrm{out}}\|_2^2
    +
    \mathbb{E}[\|a_r\|_2^2 \mid X_{\mathrm{in}}].
\]
Substituting the expression for $a_r$ gives
\[
    \mathbb{E}[\|a_r\|_2^2 \mid X_{\mathrm{in}}]
    =
    \|\beta_{\mathrm{out}}\|_2^2
    \operatorname{tr}
    \left(
        (C_u+rC_d)^{-1}
        (C_u+r^2C_d)
        (C_u+rC_d)^{-1}
    \right),
\]
and therefore
\[
    \mathbb{E}[L_{\mathrm{test}} \mid X_{\mathrm{in}}]
    =
    \|\beta_{\mathrm{out}}\|_2^2
    \left[
        1+
        \operatorname{tr}
        \left(
            (C_u+rC_d)^{-1}
            (C_u+r^2C_d)
            (C_u+rC_d)^{-1}
        \right)
    \right].
\]

\section{Theory: Repetition Peak Heatmap}
\label{app:heatmap}

\begin{figure}[t]
  \centering
  \includegraphics[width=\linewidth]{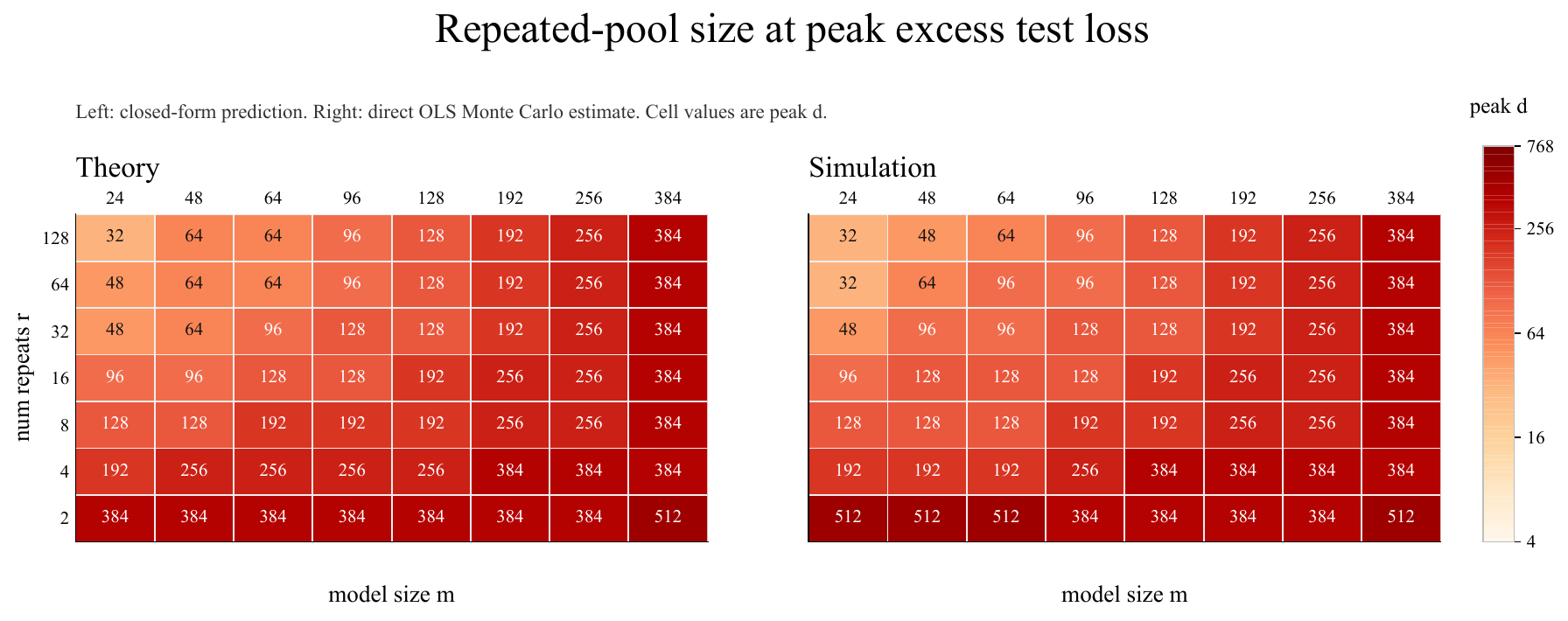}
  \caption{
  To isolate the peak location, we fix $(m,r)$ and report the repeated-pool size $d$ that maximizes excess test loss. The same trend appears in both the closed-form risk and direct OLS simulations: larger-capacity models peak at larger repeated pools, while higher repeat counts shift the peak toward smaller pools. We note that theory and simulation agree up to the resolution of the $d$ grid.
  }
  \label{fig:peak-heat}
\end{figure}

\begin{figure}[t]
  \centering
  \includegraphics[width=\linewidth]{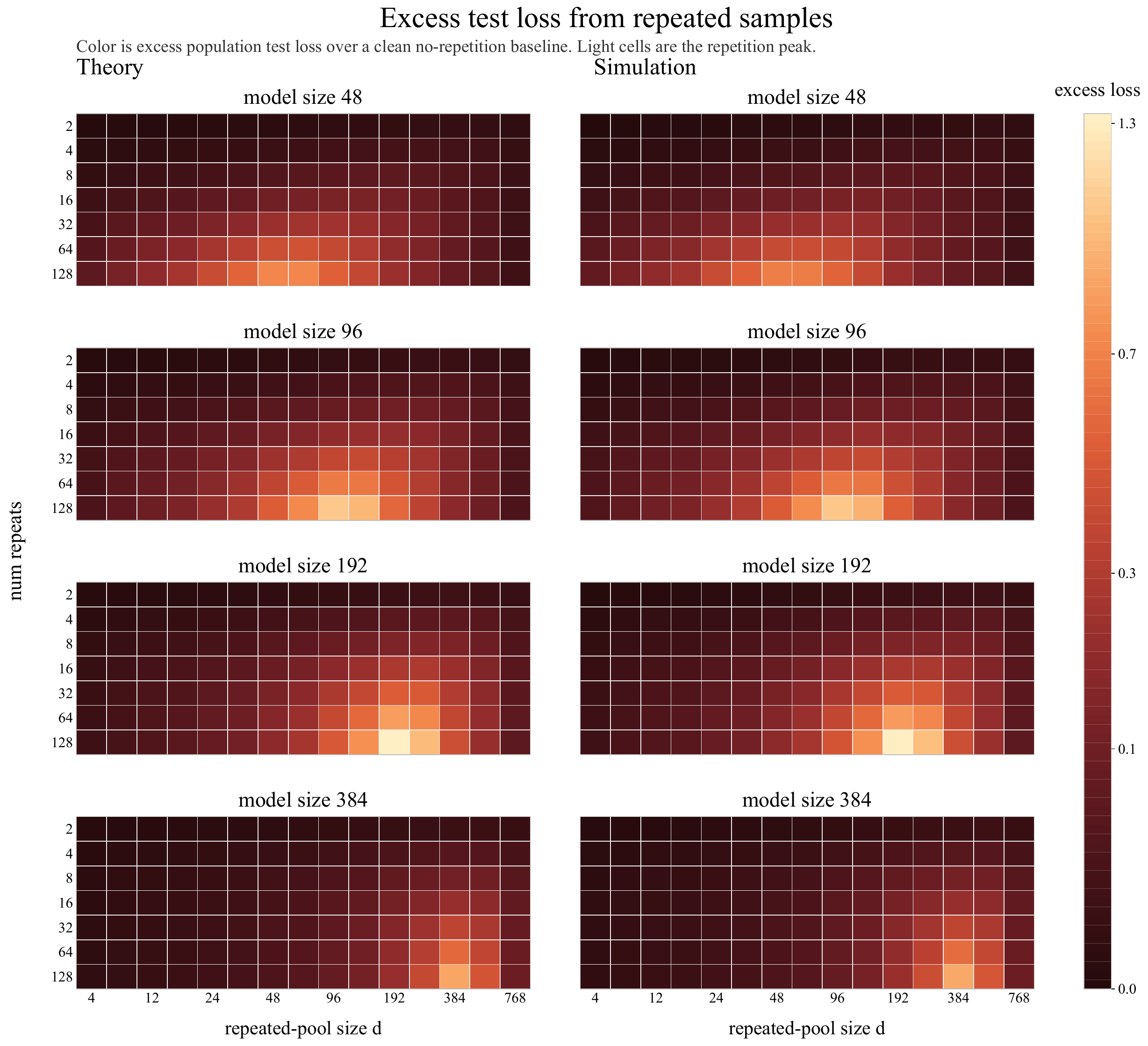}
  \caption{
  Full visualization of Figure~\ref{fig:linear-loss-surfaces}
  }
  \label{fig:full-one}
\end{figure}

\section{Training and evaluation details}
\label{app:training-details}

\paragraph{Architecture.} We use Qwen3-style decoder-only transformers~\citep{yang2025qwen3,yang2024qwen2,yang2025qwen25} with rotary position embeddings~\citep{su2024roformer}, RMSNorm, SwiGLU feed-forward layers, grouped-query attention, and the Qwen3 tokenizer. Six parameter counts are used, $N \in \{34, 48, 63, 93, 153, 344\}\mathrm{M}$, obtained by scaling depth and width approximately uniformly. We compute $C = 6NT$ FLOPs per the standard dense-transformer estimate~\citep{kaplan2020scaling,hoffmann2022training}.

\paragraph{Optimizer and schedule.}  All runs use AdamW \cite{kingma2017adammethodstochasticoptimization} with the fused PyTorch implementation (\texttt{adamw\_torch\_fused}), $\beta_1=0.9$, $\beta_2=0.95$,
weight decay $0.01$, and gradient clipping at $1.0$. The learning rate
  follows a cosine schedule with warmup ratio $0.2$. The peak learning rate
  is derived from a base learning rate of $10^{-6}$ and the computed
  optimizer-step token count, rather than tuned separately per model size.
  Sequence length is 2048 throughout. Training uses BF16 weights,
  FlashAttention-2, and torch compilation.


\paragraph{Data and evaluation.}
  The source corpus is FineWeb-Edu-Dedup from the HuggingFaceTB
  SmolLM corpus \citet{penedo2024fineweb}. We split the corpus once with train/test split seed 0,
  holding out approximately 150M tokens for evaluation before constructing
  any repeated pools. Thus evaluation documents are excluded from both the
  non-repeated training stream and the repeated pool. Documents are tokenized
  with the Qwen3 tokenizer \cite{yang2025qwen3}, truncated to length 2048, and assigned EOS tokens.  For each repeated-data run, the repeated pool is selected at document
  granularity to contribute approximately $fT/\Rep$ unique tokens. These
  documents are then replayed $\Rep$ times, combined with non-repeated
  documents contributing approximately $(1-f)T$ tokens, and shuffled into the
  final training stream. This makes the repeated-token fraction approximately
  $f=0.1$ while varying the concentration of those repeated tokens.

 \paragraph{Sweep grid.}
  We train no-repetition baselines and repeated-data sweeps for each completed
  $(N,\OT)$ cell. The completed analysis grid contains 25 cells: all five
  overtraining multipliers for 34M, 48M, 63M, and 93M; four multipliers
  through $\OT=2$ for 153M; and $\OT=1$ for 344M. Repeat counts are swept on
  an irregular, approximately logarithmic grid spanning from $\Rep=1$ to as
  high as $\Rep=20000$, subject to the repeated pool containing at least one
  document. Each completed cell is a single training run.


\paragraph{Peak fitting.}
  For each completed $(N,\OT)$ sweep, we first compute the fractional eval-loss
  increase relative to the corresponding no-repetition baseline,
  $\eta(\Rep) = (L(\Rep)-L_{\baseline})/L_{\baseline}$. We then fit a
  three-parameter Gaussian in $\log_{10}\Rep$ to the non-baseline points
  ($\Rep>1$):
  \[
      \eta(\Rep)
      =
      A \exp\!\left(
      -\frac{(\log_{10}\Rep-\log_{10}\mu)^2}{2\sigma^2}
      \right).
  \]
  The fitted peak repeat count is $\Rep^{\mathrm{peak}}=\mu$, and the
  corresponding repeated-pool size is computed from
  $\Dr^{\mathrm{peak}} = 2\OT N/\Rep^{\mathrm{peak}}$. Power-law fits in
  \eqref{eq:peak-N}--\eqref{eq:peak-C} are ordinary least-squares regressions
  in log-log space over the completed sweeps.


\paragraph{Scaling-law fitting.}
  The no-repetition scaling law $L(C)=E+KC^{-\gamma}$ in
  \S\ref{sec:finding-3} is fit on the six $\OT=1$, $\Rep=1$ baselines. We fit
  in log-loss space using nonlinear least squares: first estimating an initial
  $K$ and $\gamma$ from a log-log linear fit without a loss floor, then refitting
  $E$, $K$, and $\gamma$ jointly with $E$ free. This gives
  \eqref{eq:frontier-fit}. The fitted curve matches the six no-repetition
  baselines to within about $0.015$ nats, and we discuss sensitivity to this
  three-parameter fit in \S\ref{sec:finding-3}.  

\paragraph{Reported evaluation loss.}
  All reported eval losses are final-checkpoint losses. After each run reaches its target token budget $T = 20 \cdot \OT \cdot N$, up to document-boundary rounding from the sampling
  procedure, we evaluate the final checkpoint once on the fixed held-out split and use that value in all figures, peak fits, and scaling-law fits. We use the same rule for repeated-data
  runs and no-repetition baselines. We do not report the best validation checkpoint or an average over checkpoints; intermediate evaluations are used only for monitoring.

\section{Interpreting \texorpdfstring{$\CEG>1$}{CEG > 1}}  \label{app:cs-greater-than-one}

  The $\CEG$ ratio in \eqref{eq:ceg-cel} is defined relative to our fitted no-repetition scaling law. It is not an oracle optimum over all possible model sizes, token budgets,
  optimizers, and data mixtures. Therefore $\mathrm{CEG}>1$ can occur when a run achieves lower loss than the fitted no-repetition reference curve predicts at its actual compute. We
  interpret such values as being above this fitted reference, not as a universal claim that the run is optimally compute saving. 

  One reason this can happen is that $\OT=1$ uses the Chinchilla-style rule of 20 tokens per parameter. That rule was estimated in a different setting, with a different model family,
  optimizer stack, tokenizer, and data mixture. The optimal token-per-parameter ratio for our Qwen3-style \cite{yang2025qwen3} models on FineWeb-Edu-Dedup \cite{penedo2024fineweb} may therefore differ from 20. We do not attempt to
  find the globally optimal $(N,T)$ allocation for this model family, because the goal of the paper is to compare repetition structures at fixed budget and to convert their losses through
  a consistent no-repetition reference.

  For the repeated-data comparisons, what matters is that all runs are evaluated against the same fitted no-repetition reference, and that each repeated-data run is also compared to the
  no-repetition baseline at the same $(N,\OT)$ budget. Values below one imply positive $\CEL$ relative to this reference. Values above one imply negative $\CEL$ and should be read as a calibration effect of the fitted reference curve.

\section{Limitations}
\label{sec:limitations}

  Our experiments are small relative to frontier pretraining: the largest model has 344M parameters, and all runs use one Qwen3-style architecture family, one tokenizer, one corpus, and a
  fixed repeated-token fraction $f=0.1$. Each $(N,\OT,\Rep)$ cell is a single training run, so we do not estimate seed-level variance; the evidence for robustness comes from consistency
  across sweeps, not repeated random restarts. The larger-model grid is incomplete because of compute constraints, with 153M run through $\OT=2$ and 344M run only at $\OT=1$.  The repeat-count sweep extends to $\Rep=20000$; the log-Gaussian peak fits use the full range, though plots display only up to $\Rep \approx 3000$ for readability.
  Finally, $\CEG$ and $\CEL$ depend on a three-parameter no-repetition scaling law fit to six $\OT=1$ baselines, so their absolute values should be treated as point estimates.
  The linear-regression model explains one mechanism by which repetition can hurt, but it leaves out many features of real language models, including attention, depth, optimization
  dynamics, and discrete tokens. It should be read as an explanatory toy model, not a quantitative model of transformer pretraining.

\section{Broader Impacts}
\label{sec:broaderimpacts}

 This work studies how repeated training data can reduce the compute efficiency of language-model pretraining. Better measurement of repetition damage may help practitioners spend less
  compute and energy on runs whose data mixture is poorly structured. The same analysis could also be used to justify more aggressive data filtering, so care is needed to avoid removing
  useful minority-domain or low-resource-language data solely because it appears repetitive.

\section{Experimental compute resources}
\label{sec:computeresources}

All experiments were run on GPU clusters using BF16 training and FlashAttention-2. The main experimental grid consists of Qwen3-style models from 34M to 344M parameters, trained across
  repeated-data sweeps and no-repetition baselines as described in Appendix~\ref{app:training-details}. The largest individual runs are the 344M-parameter models at $\OT=1$, with total
  training compute estimated by $C=6NT$.


\end{document}